\documentclass[a4paper,journal,draftcls,onecolumn,12pt]{IEEEtran}%
\usepackage{amsfonts}
\usepackage{amsmath}
\usepackage{amssymb}
\usepackage{graphicx}%
\providecommand{\U}[1]{\protect\rule{.1in}{.1in}}
\begin{document}

\title{A New Continuous-Time Equality-Constrained Optimization Method to Avoid Singularity}
\author{Quan Quan and Kai-Yuan Cai\thanks{The authors are with Department of Automatic
Control, Beijing University of Aeronautics and Astronautics, Beijing 100191,
China (Quan Quan: qq\_buaa@buaa.edu.cn, http://quanquan.buaa.edu.cn/; Kai-Yuan
Cai: kycai@buaa.edu.cn)}}
\maketitle

\begin{abstract}
In equality-constrained optimization, a standard regularity assumption is
often associated with feasible point methods, namely the gradients of
constraints are linearly independent. In practice, the regularity assumption
may be violated. To avoid such a singularity, we propose a new projection
matrix, based on which a feasible point method for the continuous-time,
equality-constrained optimization problem is developed. First, the equality
constraint is transformed into a continuous-time dynamical system with
solutions that always satisfy the equality constraint. Then, the singularity
is explained in detail and a new projection matrix is proposed to avoid
singularity. An update (or say a controller) is subsequently designed to
decrease the objective function along the solutions of the transformed system.
The invariance principle is applied to analyze the behavior of the solution.
We also propose a modified approach for addressing cases in which solutions do
not satisfy the equality constraint. Finally, the proposed optimization
approaches are applied to two examples to demonstrate its effectiveness.

\end{abstract}


\section{Introduction}

According to the implementation of a differential equation, most approaches to
continuous-time optimization can be classified as either a dynamical system
\cite{Tanabe(1980)},\cite{Yamashita(1980)},\cite{Brown(1989)} or a neural
network \cite{Zhang(1992)},\cite{Hou(2001)},\cite{Liao(2004)}%
,\cite{Barbarosou(2008)}. The dynamical system approach relies on the
numerical integration of differential equations on a digital computer. Unlike
discrete optimazation methods, the step sizes of dynamical system approaches
can be controlled automatically in the integration process and can sometimes
be made larger than usual. This advantage suggests that the dynamical system
approach can in fact be comparable with currently available conventional
discrete optimal methods and facilitate faster convergence \cite{Tanabe(1980)}%
,\cite{Brown(1989)}. The application of a higher-order numerical integration
process also enables us to avoid the zigzagging phenomenon, which is often
encountered in typical linear extrapolation methods \cite{Tanabe(1980)}. On
the other hand, the neural network approach emphasizes implementation by
analog circuits, very large scale integration, and optical technologies
\cite{Absi(2005)}. The major breakthrough of this approach is attributed to
the seminal work of Hopfield, who introduced an artificial neural network to
solve the traveling salesman problem (TSP) \cite{Hopfield(1985)}. By employing
analog hardware, the neural network approach offers low computational
complexity and is suitable for parallel implementation.

For continuous-time equality-constrained optimization, existing methods can be
classified into three categories \cite{Tanabe(1980)}: feasible point method
(or primal method), augmented function method (or penalty function method),
and the Lagrangian multiplier method. Determining whether one method
outperforms the others is difficult because each method possesses distinct
advantages and disadvantages. Readers can refer to \cite{Tanabe(1980)}%
,\cite{Zhang(1992)},\cite{Barbarosou(2008)},\cite{Luenberger(2007)} and the
references therein for details. The feasible point method directly solves the
original problem by searching through the feasible region for the optimal
solution. Each point in the process is feasible, and the value of the
objective function constantly decreases. Compared with the two other methods,
the feasible point method offers three significant advantages that highlight
its usefulness as a general procedure that is applicable to almost all
nonlinear programming problems \cite[p. 360]{Luenberger(2007)}: i) the
terminating point is feasible if the process is terminated before the solution
is reached; ii) the limit point of the convergent sequence of solutions must
be at least a local constrained minimum; and iii) the approach is applicable
to general nonlinear programming problems because it does not rely on special
problem structures such as convexity.

In this paper, a continuous-time feasible point approach is proposed for
equality-constrained optimization. First, the equality constraint is
transformed into a continuous-time dynamical system with solutions that always
satisfy the equality constraint. Then, the singularity is explained in detail
and a new projection matrix is proposed to avoid singularity. An update (or
say a controller) is subsequently designed to decrease the objective function
along the solutions of the transformed system. The invariance principle is
applied to analyze the behavior of the solution. We also propose a modified
approach for addressing cases in which solutions do not satisfy the equality
constraint. Finally, the proposed optimization approach is applied to two
examples to demonstrate its effectiveness.

Local convergence results do not assume convexity in the optimization problem
to be solved. Compared with global optimization methods, local optimization
methods are still necessary. First, they often server as a basic component for
some global optimizations, such as the branch and bound method
\cite{Lawler(1966)}. On the other hand, they can require less computation for
online optimization. Compared with the discrete optimal methods offered by
MATLAB, at least two illustrative examples show that the proposed approach
avoids convergence to a singular point and facilitates faster convergence
through numerical integration on a digital computer. In view of these, the
contributions of this paper are clear and listed as follows.

i) A new projection matrix is proposed to remove a standard regularity
assumption that is often associated with feasible point methods, namely that
the gradients of constraints are linearly independent, see \cite[p.158,
Equ.(4)]{Tanabe(1980)},\cite[p.156, Equ.(2.3)]{Yamashita(1980)},\cite[p.1669,
Assumption 1]{Barbarosou(2008)}. Compared with a commonly-used modified
projection matrix, the proposed projection matrix has better precision.
Moreover, its recursive form can be implemented more easily.

ii) Based on the proposed matrix, a continuous-time, equality-constrained
optimization method is developed to avoid convergence to a singular point. The
invariance principle is applied to analyze the behavior of the solution.

iii) The modified version of the proposed optimization is further developed to
address cases in which solutions do not satisfy the equality constraint. This
ensures its robustness against uncertainties caused by numerical error or
realization by analog hardware.

We use the following notation. $%
%TCIMACRO{\U{211d} }%
%BeginExpansion
\mathbb{R}
%EndExpansion
^{n}$ is Euclidean space of dimension $n$. $\left\Vert \mathbf{\cdot
}\right\Vert $ denotes the Euclidean vector norm or induced matrix norm.
$I_{n}$ is the identity matrix with dimension $n.$ $0_{n_{1}\times n_{2}}$
denotes a zero vector or a zero matrix with dimension $n_{1}\times n_{2}.$
Direct product $\otimes$ and $\operatorname{vec}\left(  \cdot\right)
$\ operation are defined in \textit{Appendix A}. The function $\left[
\cdot\right]  _{\times}:%
%TCIMACRO{\U{211d} }%
%BeginExpansion
\mathbb{R}
%EndExpansion
^{3}$ $\rightarrow$ $%
%TCIMACRO{\U{211d} }%
%BeginExpansion
\mathbb{R}
%EndExpansion
^{3\times3}\ $with matrix $H\in%
%TCIMACRO{\U{211d} }%
%BeginExpansion
\mathbb{R}
%EndExpansion
^{9\times3}$ is defined in \textit{Appendix B}. Suppose $g:$ $%
%TCIMACRO{\U{211d} }%
%BeginExpansion
\mathbb{R}
%EndExpansion
^{n}\rightarrow%
%TCIMACRO{\U{211d} }%
%BeginExpansion
\mathbb{R}
%EndExpansion
.$ The gradient of the function $g$ is given by $\nabla g\left(  x\right)
=\nabla_{x}g\left(  x\right)  =[\partial g\left(  x\right)  \left/  \partial
x_{1}\right.  $ $\cdots\ \partial g\left(  x\right)  \left/  \partial
x_{n}\right.  ]^{T}\in%
%TCIMACRO{\U{211d} }%
%BeginExpansion
\mathbb{R}
%EndExpansion
^{n}\ $and the matrix of second partial derivatives of $g\left(  x\right)  $
known as Hessian is given by $\nabla_{xx}$ $:$ $%
%TCIMACRO{\U{211d} }%
%BeginExpansion
\mathbb{R}
%EndExpansion
$ $\rightarrow$ $%
%TCIMACRO{\U{211d} }%
%BeginExpansion
\mathbb{R}
%EndExpansion
^{n\times n}$ and $\nabla_{xx}g\left(  x\right)  =\left[  \partial^{2}g\left(
x\right)  \left/  \partial x_{i}\partial x_{j}\right.  \right]  _{ij}.$

\section{Problem Formulation}

\subsection{Equality-Constrained Optimization}

The class of equality-constrained optimization problems considered here is
defined as follows:%
\begin{equation}
\underset{x\in%
%TCIMACRO{\U{211d} }%
%BeginExpansion
\mathbb{R}
%EndExpansion
^{n}}{\min}v\left(  x\right)  ,\text{ s.t. }c\left(  x\right)  =0
\label{optimazation}%
\end{equation}
where $v:$ $%
%TCIMACRO{\U{211d} }%
%BeginExpansion
\mathbb{R}
%EndExpansion
^{n}\rightarrow%
%TCIMACRO{\U{211d} }%
%BeginExpansion
\mathbb{R}
%EndExpansion
$ is the objective function and $c=[c_{1}$ $c_{2}$ $\cdots$ $c_{m}]^{T}\in%
%TCIMACRO{\U{211d} }%
%BeginExpansion
\mathbb{R}
%EndExpansion
^{m},$ $c_{i}:%
%TCIMACRO{\U{211d} }%
%BeginExpansion
\mathbb{R}
%EndExpansion
^{n}\rightarrow%
%TCIMACRO{\U{211d} }%
%BeginExpansion
\mathbb{R}
%EndExpansion
$ are the equality constraints. They are both twice continuously
differentiable. Denote by $\nabla c\left(  x\right)  \triangleq\left[
\begin{array}
[c]{cccc}%
\nabla c_{1}\left(  x\right)  & \nabla c_{2}\left(  x\right)  & \cdots &
\nabla c_{m}\left(  x\right)
\end{array}
\right]  \in%
%TCIMACRO{\U{211d} }%
%BeginExpansion
\mathbb{R}
%EndExpansion
^{n\times m}.$ To avoid a trivial case, suppose the constraint (or feasible
set) $\mathcal{F=}\left\{  \left.  x\in%
%TCIMACRO{\U{211d} }%
%BeginExpansion
\mathbb{R}
%EndExpansion
^{n}\right\vert c\left(  x\right)  =0\right\}  \mathcal{\neq\emptyset}$.

\textbf{Definition 1 }\cite[pp. 316-317]{Jorge(1999)}. For the problem
(\ref{optimazation}), a vector $x^{\ast}\in\mathcal{F}$ is a global minimum if
$v\left(  x^{\ast}\right)  \leq v\left(  x\right)  ,$ $\forall x\in
\mathcal{F};$ a vector $x^{\ast}\in\mathcal{F}$ is a local (strict local)
minimum if there is a neighborhood $\mathcal{N}$ of $x^{\ast}$ such that
$v\left(  x^{\ast}\right)  \leq v\left(  x\right)  \ (v\left(  x^{\ast
}\right)  <v\left(  x\right)  )$ for $x\in\mathcal{N\cap F}.$

\textbf{Definition 2 }\cite[p. 325]{Luenberger(2007)}. A vector $x^{\ast}%
\in\mathcal{F}$ is said to be a regular point if the gradient vectors $\nabla
c_{1}\left(  x^{\ast}\right)  ,\nabla c_{2}\left(  x^{\ast}\right)
,\cdots,\nabla c_{m}\left(  x^{\ast}\right)  $ are linearly independent.
Otherwise, it is called a singular point.

This paper aims to propose an approach to continuous-time,
equality-constrained optimization to identify the local minima based on a
feedback control perspective.

\textbf{Remark 1. }Inequality-constrained optimizations can be transformed
into equality-constrained optimizations by introducing new variables. For
example, the inequality constraint $x\leq1,x\in%
%TCIMACRO{\U{211d} }%
%BeginExpansion
\mathbb{R}
%EndExpansion
$ can be replaced with an equality constraint $x+z^{2}=1,z\in%
%TCIMACRO{\U{211d} }%
%BeginExpansion
\mathbb{R}
%EndExpansion
.$ Also, the inequality constraint $-1\leq x\leq1,x\in%
%TCIMACRO{\U{211d} }%
%BeginExpansion
\mathbb{R}
%EndExpansion
$ can be replaced with an equality constraint $x=\sin\left(  z\right)  ,z\in%
%TCIMACRO{\U{211d} }%
%BeginExpansion
\mathbb{R}
%EndExpansion
.$ Here, we only focus on equality-constrained optimization.

\subsection{Equality Constraint Transformation}

Optimization problems are often solved by using numerical iterative methods.
For an equality-constrained optimization problem, the major difficulty lies in
ensuring that each iteration satisfies the constraint and can further move
toward the minimum. To address this difficulty, a transformation of the
equality constraint is proposed, which is formulated as an assumption.

\textbf{Assumption 1}.\textbf{ }For a given $x_{0}\in\mathcal{F},$ there
exists a function $f:%
%TCIMACRO{\U{211d} }%
%BeginExpansion
\mathbb{R}
%EndExpansion
^{n}\rightarrow%
%TCIMACRO{\U{211d} }%
%BeginExpansion
\mathbb{R}
%EndExpansion
^{n\times l}$ such that%
\begin{equation}
\dot{x}\left(  t\right)  =f\left(  x\left(  t\right)  \right)  u\left(
t\right)  ,x\left(  0\right)  =x_{0} \label{dynamical system}%
\end{equation}
with solutions that satisfy $x\left(  t\right)  \in\mathcal{F}_{u}\left(
x_{0}\right)  ,$ where $\mathcal{F}_{u}\left(  x_{0}\right)  =\{x\left(
t\right)  \in\mathcal{F}|\dot{x}\left(  t\right)  =f\left(  x\left(  t\right)
\right)  u\left(  t\right)  ,$ $x\left(  0\right)  =x_{0}\in\mathcal{F},$
$\forall u\left(  t\right)  \in%
%TCIMACRO{\U{211d} }%
%BeginExpansion
\mathbb{R}
%EndExpansion
^{l},$ $t\geq0\}$.

From a feedback control perspective, the update $u$ can be considered as a
control input. The objective function $v\left(  x\right)  $ can be considered
a Lyapunov-like function, although $v(x)$ is not required to be a Lyapunov
function. Based on \textit{Assumption 1}, the objective of this paper can be
restated as: to design a control input $u$ to decrease $v(x)$ along the
solutions of (\ref{dynamical system}) until $x$ has achieved a local minimum.
In the following, we will omit the variable $t$ except when necessary.

\textbf{Remark 2}. The proposition of \textit{Assumption 1} is motivated by
the property of attitude kinematics \cite[p. 200]{Alberto Isdori (2003)}:
$\dot{x}=\frac{1}{2}E\left(  x\right)  w$, where $x=[q_{0}\ q^{T}]^{T}\in%
%TCIMACRO{\U{211d} }%
%BeginExpansion
\mathbb{R}
%EndExpansion
^{4},$ $q_{0}\in%
%TCIMACRO{\U{211d} }%
%BeginExpansion
\mathbb{R}
%EndExpansion
,\ q,w\in%
%TCIMACRO{\U{211d} }%
%BeginExpansion
\mathbb{R}
%EndExpansion
^{3}$ and $E\left(  x\right)  =[-q$ $q_{0}I_{3}+[q]_{\times}^{T}]^{T}\in%
%TCIMACRO{\U{211d} }%
%BeginExpansion
\mathbb{R}
%EndExpansion
^{4\times3}.$ The function $\left[  \cdot\right]  _{\times}:%
%TCIMACRO{\U{211d} }%
%BeginExpansion
\mathbb{R}
%EndExpansion
^{3}$ $\rightarrow$ $%
%TCIMACRO{\U{211d} }%
%BeginExpansion
\mathbb{R}
%EndExpansion
^{3\times3}$ is defined in \textit{Appendix B. }All solutions of the attitude
kinematics satisfy the constraint $\left\Vert x\right\Vert ^{2}=1$ driven by
any $w\in%
%TCIMACRO{\U{211d} }%
%BeginExpansion
\mathbb{R}
%EndExpansion
^{3}$. The explanation is given as follows. It is easy to check that
$x^{T}\dot{x}=\frac{1}{2}x^{T}E\left(  x\right)  w=0\ $since $[q]_{\times}q=0$
for$\ \forall q\in%
%TCIMACRO{\U{211d} }%
%BeginExpansion
\mathbb{R}
%EndExpansion
^{3}.$ Therefore, the solution always satisfies the constraint $\left\Vert
x\left(  t\right)  \right\Vert ^{2}=1$ if $\left\Vert x\left(  0\right)
\right\Vert =1,$ $t\geq0.$ Another representation of attitude kinematics is
\begin{equation}
\dot{R}=\left[  w\right]  _{\times}R \label{rotation}%
\end{equation}
where $R\in%
%TCIMACRO{\U{211d} }%
%BeginExpansion
\mathbb{R}
%EndExpansion
^{3\times3}$ is a rotation matrix satisfying the constraint $R^{T}R=I_{3}$.
For (\ref{rotation}), we have%
\begin{align*}
\frac{d}{dt}\left(  R^{T}R\right)   &  =R^{T}\dot{R}+\dot{R}^{T}R\\
&  =R^{T}\left(  \left[  w\right]  _{\times}+\left[  w\right]  _{\times}%
^{T}\right)  R=0_{3\times3}.
\end{align*}
That is why the evolution of $R$ always lies on the constraint $R^{T}R=I_{3}.$

\textbf{Remark 3}. The best choice of $f\left(  x\right)  $ is to satisfy
$\mathcal{F}_{u}\left(  x_{0}\right)  =\mathcal{F}.$ However, it is difficult
to achieve. For example, if $c\left(  x\right)  =\left(  x_{1}+1\right)
\left(  x_{1}-1\right)  $, $x=[x_{1}$ $x_{2}]^{T}\in%
%TCIMACRO{\U{211d} }%
%BeginExpansion
\mathbb{R}
%EndExpansion
^{2},$ then $\mathcal{F}=\left\{  \left.  x\in%
%TCIMACRO{\U{211d} }%
%BeginExpansion
\mathbb{R}
%EndExpansion
^{2}\right\vert x_{1}=1,x_{1}=-1\right\}  $. Since the two sets $\left\{
\left.  x\in%
%TCIMACRO{\U{211d} }%
%BeginExpansion
\mathbb{R}
%EndExpansion
^{2}\right\vert x_{1}=1\right\}  $ and $\left\{  \left.  x\in%
%TCIMACRO{\U{211d} }%
%BeginExpansion
\mathbb{R}
%EndExpansion
^{2}\right\vert x_{1}=-1\right\}  $ are not connected, the solution of
(\ref{dynamical system}) starting from either set cannot access the other.
Although $\mathcal{F}_{u}\left(  x_{0}\right)  \neq\mathcal{F}$, we still
expect the global minimum $x^{\ast}\in\mathcal{F}_{u}\left(  x_{0}\right)  .$
That is why we often require that the initial value $x_{0}$ be close to the
global minimum $x^{\ast}.$ Besides this, it is also expected that the function
$f\left(  x\right)  $ is chosen to make the set $\mathcal{F}_{u}\left(
x_{0}\right)  $\ as large as possible so that the probability of $x^{\ast}%
\in\mathcal{F}_{u}\left(  x_{0}\right)  $ is higher.

If $c\left(  x\right)  =Ax,$ $A\in%
%TCIMACRO{\U{211d} }%
%BeginExpansion
\mathbb{R}
%EndExpansion
^{m\times n}$, then the function $f\left(  x\right)  $ can be chosen to
satisfy $\mathcal{F}=\mathcal{F}_{u}\left(  x_{0}\right)  ,$ $\forall x_{0}%
\in\mathcal{F}.$

\textbf{Theorem 1}. Suppose that $c\left(  x\right)  =Ax$ and $f\left(
x\right)  =A^{\bot},$ where $A^{\bot}$ is with full column rank, and the space
spanned by the columns of $A^{\bot}$ is the null space of $A.$ Then
$\mathcal{F}=\mathcal{F}_{u}\left(  x_{0}\right)  ,$ $\forall x_{0}%
\in\mathcal{F}.$

\textit{Proof.} Since $\mathcal{F}_{u}\left(  x_{0}\right)  \subseteq
\mathcal{F},$ the remaining task is to prove $\mathcal{F}\subseteq
\mathcal{F}_{u}\left(  x_{0}\right)  ,$ $\forall x_{0}\in\mathcal{F},$\ namely
for any $\bar{x}\in\mathcal{F}$ there exists a control input $u\in%
%TCIMACRO{\U{211d} }%
%BeginExpansion
\mathbb{R}
%EndExpansion
^{l}$ that can transfer any initial state $x_{0}\in\mathcal{F}$ to $\bar{x}.$
Since $x_{0},\bar{x}\in\mathcal{F},$ there exist $u_{0},\bar{u}\in%
%TCIMACRO{\U{211d} }%
%BeginExpansion
\mathbb{R}
%EndExpansion
^{l}$ such that $\bar{x}=A^{\bot}\bar{u}$ and $x\left(  0\right)  =A^{\bot
}u_{0}$ by the definition of $A^{\bot}.$ Design a control input
\[
u\left(  t\right)  =\left\{
\begin{array}
[c]{c}%
\frac{1}{\bar{t}}\left(  \bar{u}-u_{0}\right)  ,\\
0,
\end{array}%
\begin{array}
[c]{c}%
0\leq t\leq\bar{t}\\
t>\bar{t}.
\end{array}
\right.  .
\]
With the control input above, we have
\begin{align*}
x\left(  t\right)  -x\left(  0\right)   &  =%
%TCIMACRO{\dint \nolimits_{0}^{t}}%
%BeginExpansion
{\displaystyle\int\nolimits_{0}^{t}}
%EndExpansion
A^{\bot}u\left(  s\right)  ds\\
&  =%
%TCIMACRO{\dint \nolimits_{0}^{\bar{t}}}%
%BeginExpansion
{\displaystyle\int\nolimits_{0}^{\bar{t}}}
%EndExpansion
A^{\bot}u\left(  s\right)  ds=A^{\bot}\bar{u}-A^{\bot}u_{0},
\end{align*}
when $t\geq\bar{t}$. Then $x\left(  t\right)  =\bar{x},$ $t\geq\bar{t}.$ Hence
$\mathcal{F}\subseteq\mathcal{F}_{u}\left(  x_{0}\right)  ,$ $\forall x_{0}%
\in\mathcal{F}.$ Consequently, $\mathcal{F}=\mathcal{F}_{u}\left(
x_{0}\right)  ,$ $\forall x_{0}\in\mathcal{F}.$ $\square$

From the proof of \textit{Theorem 1}, the choice of $f\left(  x\right)  $
becomes a controllability problem. However, it is difficult to obtain a
controllability condition of a general nonlinear system. Correspondingly, it
is difficult to choose $f\left(  x\right)  $ for a general nonlinear function
$c\left(  x\right)  $ to satisfy $\mathcal{F}=\mathcal{F}_{u}\left(
x_{0}\right)  .$ Motivated by the linear case above, we aim to design a
function $f\left(  x\right)  \ $whose range is the null space of $\nabla
c\left(  x\right)  ^{T}$ for any fixed $x\in%
%TCIMACRO{\U{211d} }%
%BeginExpansion
\mathbb{R}
%EndExpansion
^{n}.$ This idea can be formulated as $\mathcal{V}_{1}\left(  x\right)
=\mathcal{V}_{2}\left(  x\right)  $, where
\begin{align*}
\mathcal{V}_{1}\left(  x\right)   &  =\{z\in%
%TCIMACRO{\U{211d} }%
%BeginExpansion
\mathbb{R}
%EndExpansion
^{n}|\nabla c\left(  x\right)  ^{T}z=0\},\\
\mathcal{V}_{2}\left(  x\right)   &  =\{z\in%
%TCIMACRO{\U{211d} }%
%BeginExpansion
\mathbb{R}
%EndExpansion
^{n}|z=f\left(  x\right)  u,u\in%
%TCIMACRO{\U{211d} }%
%BeginExpansion
\mathbb{R}
%EndExpansion
^{l}\}.
\end{align*}

\section{Singularity and A New Projection Matrix}

\subsection{Singularity}

The function $f$ is the projection matrix, which orthogonally projects a
vector onto the null space of $\nabla c^{T}$. One well-known projection matrix
is given as follows \cite{Tanabe(1980)},\cite{Yamashita(1980)}%
,\cite{Barbarosou(2008)}:%
\begin{equation}
f\left(  x\right)  =I_{n}-\left(  \nabla c\left(  \nabla c^{T}\nabla c\right)
^{-1}\nabla c^{T}\right)  \left(  x\right)  . \label{projectionmatrix}%
\end{equation}
We can easily verify that $\nabla c\left(  x\right)  ^{T}f\left(  x\right)
\equiv0.$ This projection matrix requires that $\nabla c\left(  x\right)  $
should have full column rank, i.e., every $x\in\mathcal{F}$ is a regular
point. However, the assumption does not hold in cases where $\nabla c\left(
x\right)  ^{T}\nabla c\left(  x\right)  $ is singular. This condition is the
major motivation of this paper. For example, consider an equality constraint
as%
\[
c\left(  x\right)  =\left(  x_{1}-x_{2}+2\right)  \left(  x_{1}+x_{2}\right)
=0,
\]
where $x=\left[
\begin{array}
[c]{cc}%
x_{1} & x_{2}%
\end{array}
\right]  ^{T}\in%
%TCIMACRO{\U{211d} }%
%BeginExpansion
\mathbb{R}
%EndExpansion
^{2}.$ The feasible set is either $\left\{  \left.  x\in%
%TCIMACRO{\U{211d} }%
%BeginExpansion
\mathbb{R}
%EndExpansion
^{2}\right\vert x_{1}-x_{2}+2=0\right\}  $ or $\left\{  \left.  x\in%
%TCIMACRO{\U{211d} }%
%BeginExpansion
\mathbb{R}
%EndExpansion
^{2}\right\vert x_{1}+x_{2}=0\right\}  .$ As shown in Fig.1, the point
$x_{p_{1}}=\left[
\begin{array}
[c]{cc}%
-2 & 0
\end{array}
\right]  ^{T}$ has a unique feasible direction and the point $x_{p_{2}%
}=\left[
\begin{array}
[c]{cc}%
0 & 0
\end{array}
\right]  ^{T}$ also has a unique feasible direction. Whereas, the point
$x_{p_{3}}=\left[
\begin{array}
[c]{cc}%
-1 & 1
\end{array}
\right]  ^{T}$ has two feasible directions. This causes the singular
phenomena. The singularity often occurs at the intersection of the feasible
sets, where exist non-unique feasible directions. Mathematically, $\nabla
c\left(  x\right)  ^{T}\nabla c\left(  x\right)  $ is singular. Concretely,
the gradient vector of $c\left(  x\right)  $ is%
\[
\nabla c\left(  x\right)  =\left[
\begin{array}
[c]{c}%
2x_{1}+2\\
-2x_{2}+2
\end{array}
\right]  .
\]
At the points $x_{p_{1}}$ and $x_{p_{2}},$ the gradient vector of $c\left(
x\right)  $ is%
\[
\nabla c\left(  x_{p_{1}}\right)  =\left[
\begin{array}
[c]{c}%
-2\\
2
\end{array}
\right]  ,\nabla c\left(  x_{p_{2}}\right)  =\left[
\begin{array}
[c]{c}%
2\\
2
\end{array}
\right]
\]
and by (\ref{projectionmatrix}), the projection matrices are further
\[
f\left(  x_{p_{1}}\right)  =\left[
\begin{array}
[c]{cc}%
0 & 1\\
1 & 0
\end{array}
\right]  ,f\left(  x_{p_{2}}\right)  =\left[
\begin{array}
[c]{cc}%
0 & -1\\
-1 & 0
\end{array}
\right]
\]
respectively. Whereas, at the point $x_{p_{3}},$ the gradient vector of
$c\left(  x_{p_{3}}\right)  $ is%
\[
\nabla c\left(  x_{p_{3}}\right)  =\left[
\begin{array}
[c]{c}%
0\\
0
\end{array}
\right]  .
\]
For such a case, $\left(  \nabla c\left(  x_{p_{3}}\right)  ^{T}\nabla
c\left(  x_{p_{3}}\right)  \right)  ^{-1}$ does not exist.\begin{figure}[h]
\begin{center}
\includegraphics[
scale= 0.8 ]{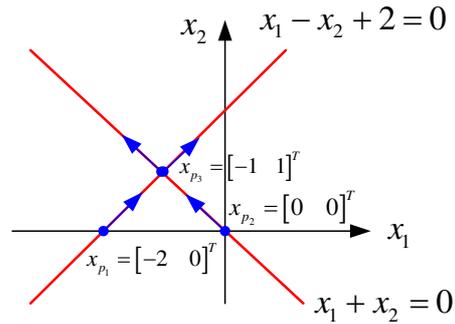}
\end{center}
\caption{Singularity Example}%
\end{figure}

To avoid singularity, a commonly-used modified projection matrix is given as
follows%
\begin{equation}
f\left(  x\right)  =I_{n}-\left(  \nabla c\left(  \varepsilon I_{m}+\nabla
c^{T}\nabla c\right)  ^{-1}\nabla c^{T}\right)  \left(  x\right)
\label{modifiedprojectionmatrix}%
\end{equation}
where $\varepsilon>0$ is a small positive scale. We have $\nabla c\left(
x\right)  ^{T}f\left(  x\right)  \neq0$ no matter how small $\varepsilon$ is.
On the other hand, to obtain $f\left(  x\right)  $ by
(\ref{modifiedprojectionmatrix}), a very small $\varepsilon$ will cause
ill-conditioning problem especially for a low-precision processor. For
example, consider the following gradient vectors:%
\begin{align}
\nabla c_{1}  &  =\left[
\begin{array}
[c]{cccc}%
1 & 1 & 1 & 1
\end{array}
\right] \nonumber\\
\nabla c_{2}  &  =\left[
\begin{array}
[c]{cccc}%
2 & 1 & 1 & 1
\end{array}
\right] \nonumber\\
\nabla c_{3}  &  =\left[
\begin{array}
[c]{cccc}%
3 & 2 & 2 & 2
\end{array}
\right]  . \label{Gradient}%
\end{align}
Taking $e_{p}=\left\Vert \nabla c^{T}f\right\Vert $ as the precision error, we
employ (\ref{modifiedprojectionmatrix}) with different $\varepsilon
=10^{-k},k=1,\cdots,15$ to obtain the projection matrix $f$. As shown in
Fig.2, the error varies with different $k$. The best precision error can be
achieved only at $\varepsilon=10^{-8}$ with a precision error around $10^{-8}%
$. Reducing $\varepsilon$ further will increase the numerical
error.\begin{figure}[h]
\begin{center}
\includegraphics[
scale= 0.5 ]{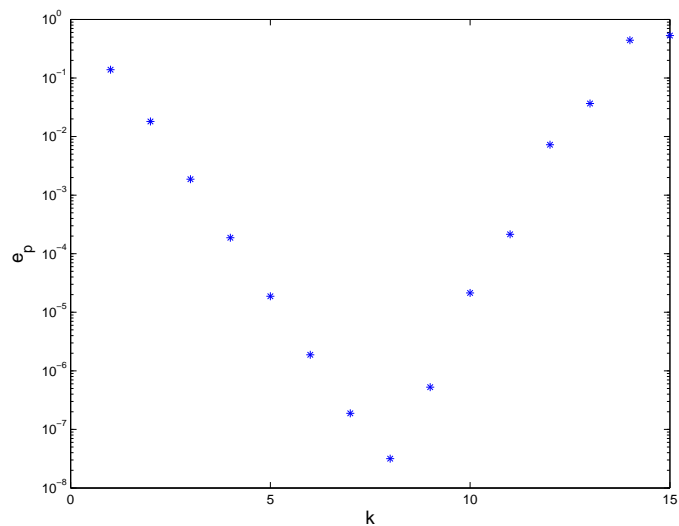}
\end{center}
\caption{Precision error of a common-used modified projection matrix with
different $\varepsilon=10^{-k}$}%
\end{figure}

The best cure is to remove the linearly dependent vector directly from $\nabla
c\left(  x\right)  $. For example, in $\nabla c\left(  x\right)  =\left[
\begin{array}
[c]{ccc}%
\nabla c_{1}\left(  x\right)  & \nabla c_{2}\left(  x\right)  & \nabla
c_{3}\left(  x\right)
\end{array}
\right]  \in%
%TCIMACRO{\U{211d} }%
%BeginExpansion
\mathbb{R}
%EndExpansion
^{n\times3}$, if $\nabla c_{3}\left(  x\right)  $ can be represented by a
linear combination of $\nabla c_{1}\left(  x\right)  \ $and $\nabla
c_{2}\left(  x\right)  ,$ then $\nabla c\left(  x\right)  ^{T}\nabla c\left(
x\right)  $ is singular. The best cure is to remove $\nabla c_{3}\left(
x\right)  $ from $\nabla c\left(  x\right)  $, resulting in%
\[
\nabla c_{new}\left(  x\right)  =\left[
\begin{array}
[c]{cc}%
\nabla c_{1}\left(  x\right)  & \nabla c_{2}\left(  x\right)
\end{array}
\right]  \in%
%TCIMACRO{\U{211d} }%
%BeginExpansion
\mathbb{R}
%EndExpansion
^{n\times2}.
\]
With it, the projection matrix becomes%
\[
f_{new}\left(  x\right)  =I_{n}-\left(  \nabla c_{new}\left(  \nabla
c_{new}^{T}\nabla c_{new}\right)  ^{-1}\nabla c_{new}^{T}\right)  \left(
x\right)  .
\]
It is easy to see that $\nabla c\left(  x\right)  ^{T}f_{new}\left(  x\right)
\equiv0.$ For a linear time-invariant matrix $\nabla c\left(  x\right)  ,$
namely independent of $x$, we can avoid singularity by removing dependent
terms out of $\nabla c\left(  x\right)  $ before computing a projection
matrix. However, this idea does not work for a general $\nabla c\left(
x\right)  $ depending on $x.$ Therefore, \textquotedblleft the best
cure\textquotedblright\ cannot be implemented continuously, which further
cannot be realized by analog hardware. For such a purpose, we will propose a
new projection matrix.

\subsection{A New Projection Matrix}

For a special case $c:%
%TCIMACRO{\U{211d} }%
%BeginExpansion
\mathbb{R}
%EndExpansion
^{n}\rightarrow%
%TCIMACRO{\U{211d} }%
%BeginExpansion
\mathbb{R}
%EndExpansion
,$ such a $f\left(  x\right)  $ is designed in \textit{Theorem 2}.
Consequently, a method is proposed to construct a projection matrix for a
general case $c:%
%TCIMACRO{\U{211d} }%
%BeginExpansion
\mathbb{R}
%EndExpansion
^{n}\rightarrow%
%TCIMACRO{\U{211d} }%
%BeginExpansion
\mathbb{R}
%EndExpansion
^{m}$. Before the design, we have the following preliminary results.

\textbf{Lemma 1}. Let
\begin{align*}
\mathcal{W}_{1}  &  =\{z\in%
%TCIMACRO{\U{211d} }%
%BeginExpansion
\mathbb{R}
%EndExpansion
^{n}|L^{T}z=0\}\\
\mathcal{W}_{2}  &  =\{z\in%
%TCIMACRO{\U{211d} }%
%BeginExpansion
\mathbb{R}
%EndExpansion
^{n}|z=\left(  I_{n}-\frac{LL^{T}}{\delta\left(  \left\Vert L\right\Vert
^{2}\right)  +\left\Vert L\right\Vert ^{2}}\right)  u,u\in%
%TCIMACRO{\U{211d} }%
%BeginExpansion
\mathbb{R}
%EndExpansion
^{n}\},
\end{align*}
where $L\in%
%TCIMACRO{\U{211d} }%
%BeginExpansion
\mathbb{R}
%EndExpansion
^{n}$ and $\delta\left(  x\right)  =\left\{
\begin{array}
[c]{c}%
1\\
0
\end{array}
\right.
\begin{array}
[c]{c}%
x=0,x\in%
%TCIMACRO{\U{211d}}%
%BeginExpansion
\mathbb{R}%
%EndExpansion
\\
x\neq0,x\in%
%TCIMACRO{\U{211d}}%
%BeginExpansion
\mathbb{R}%
%EndExpansion
\end{array}
.$ Then $\mathcal{W}_{1}=\mathcal{W}_{2}.$

\textit{Proof.} See \textit{Appendix C}. $\square$

\textbf{Theorem 2}. Suppose that $c:%
%TCIMACRO{\U{211d} }%
%BeginExpansion
\mathbb{R}
%EndExpansion
^{n}\rightarrow%
%TCIMACRO{\U{211d} }%
%BeginExpansion
\mathbb{R}
%EndExpansion
$ and the function $f\left(  x\right)  $ is designed to be%
\begin{equation}
f\left(  x\right)  =I_{n}-\frac{\nabla c\left(  x\right)  \nabla c\left(
x\right)  ^{T}}{\delta\left(  \left\Vert \nabla c\left(  x\right)  \right\Vert
^{2}\right)  +\left\Vert \nabla c\left(  x\right)  \right\Vert ^{2}}.
\label{fx}%
\end{equation}
Then \textit{Assumption 1 }is satisfied with $u\in%
%TCIMACRO{\U{211d} }%
%BeginExpansion
\mathbb{R}
%EndExpansion
^{n}$ and $\mathcal{V}_{1}\left(  x\right)  =\mathcal{V}_{2}\left(  x\right)
.$

\textit{Proof.} Since $\dot{c}\left(  x\right)  =\nabla c\left(  x\right)
^{T}\dot{x}\ $and$\ \dot{x}=f\left(  x\right)  u$, the function $f\left(
x\right)  $ is defined as in (\ref{fx}) so that $\dot{c}\left(  x\right)
\equiv0$ by \textit{Lemma 1. }Therefore,\textit{ Assumption 1 }is satisfied
with $u\in%
%TCIMACRO{\U{211d} }%
%BeginExpansion
\mathbb{R}
%EndExpansion
^{n}.$ Further by \textit{Lemma 1}, $\mathcal{V}_{1}\left(  x\right)
=\mathcal{V}_{2}\left(  x\right)  .$ $\square$

\textbf{Theorem 3}. Suppose that $c:%
%TCIMACRO{\U{211d} }%
%BeginExpansion
\mathbb{R}
%EndExpansion
^{n}\rightarrow%
%TCIMACRO{\U{211d} }%
%BeginExpansion
\mathbb{R}
%EndExpansion
^{m}$ and the function $f\left(  x\right)  $ is in a recursive form as
follows:%
\begin{align}
f_{0}  &  =I_{n}\nonumber\\
f_{k}  &  =f_{k-1}\left(  I_{n}-\frac{f_{k-1}^{T}\nabla c_{k}\nabla c_{k}%
^{T}f_{k-1}}{\delta\left(  \left\Vert f_{k-1}^{T}\nabla c_{k}\right\Vert
^{2}\right)  +\left\Vert f_{k-1}^{T}\nabla c_{k}\right\Vert ^{2}}\right)  ,
\label{modrecursiveform}%
\end{align}
\bigskip$k=1,\cdots,m.$ Then \textit{Assumption 1 }is satisfied with $f=f_{m}$
and $u\in%
%TCIMACRO{\U{211d} }%
%BeginExpansion
\mathbb{R}
%EndExpansion
^{n}$ and $\mathcal{V}_{1}\left(  x\right)  =\mathcal{V}_{2}\left(  x\right)
.$

\textit{Proof.} See \textit{Appendix D}. $\square$

\textbf{Remark 4}. In (\ref{modrecursiveform}), if $\left\Vert f_{k-1}%
^{T}\nabla c_{k}\right\Vert \neq0,$ then $\delta\left(  \left\Vert f_{k-1}%
^{T}\nabla c_{k}\right\Vert ^{2}\right)  =0,$ namely%
\[
f_{k}=f_{k-1}\left(  I_{n}-\frac{f_{k-1}^{T}\nabla c_{k}\nabla c_{k}%
^{T}f_{k-1}}{\left\Vert f_{k-1}^{T}\nabla c_{k}\right\Vert ^{2}}\right)  .
\]
This is the normal way to construct a projection matrix. On the other hand, if
$\nabla c_{k}$ can be represented by a linear combination of $\nabla c_{i},$
then $f_{k-1}^{T}\nabla c_{k}=0\ $as $f_{k-1}^{T}\nabla c_{i}=0,i=1,\cdots
,k-1.$ In this case, $\delta\left(  \left\Vert f_{k-1}^{T}\nabla
c_{k}\right\Vert ^{2}\right)  \neq0.$ Consequently, the projection matrix will
reduce to the previous one $f_{k}=f_{k-1}$, that is equivalent to removing the
term $\nabla c_{k}.$ This is consistent with \textquotedblleft the best
way\textquotedblright.

\textbf{Remark 5}. In practice, the impulse function $\delta\left(  x\right)
$ is approximated by some continuous functions such as $\delta\left(
x\right)  \approx e^{-\gamma\left\vert x\right\vert }$, where $\gamma$ is a
large positive scale. Let us revisit the example for the gradient vectors
(\ref{Gradient}). Taking $e_{p}=\left\Vert \nabla c^{T}f\right\Vert $ as the
error again, we employ (\ref{modrecursiveform}) with $\gamma=30$ to obtain the
projection matrix $f$ with $e_{p}=2.7629\ast10^{-10}.$ This demonstrates the
advantage of our proposed projection matrix over
(\ref{modifiedprojectionmatrix}). Furthermore, compared with
(\ref{projectionmatrix}) or (\ref{modifiedprojectionmatrix}), the explicit
recursive form of the proposed projection matrix is also easier for the
designer to implement.

\section{Update Design and Convergence Analysis}

In this section, by using Lyapunov's method, the update (or say controller)
$u$ is designed to result in $\dot{v}\left(  x\right)  \leq0$. However, the
objective function $v\left(  x\right)  $ is not required to be positive
definite. We base our analysis upon the LaSalle invariance theorem \cite[pp.
126-129]{Khalil (2002)}.

\subsection{Controller Design}

Taking the time derivative of $v\left(  x\right)  $ along the solutions of
(\ref{dynamical system}) results in%
\begin{equation}
\dot{v}\left(  x\right)  =\nabla v\left(  x\right)  ^{T}f\left(  x\right)
u\label{dV}%
\end{equation}
where $\nabla v\left(  x\right)  \in%
%TCIMACRO{\U{211d} }%
%BeginExpansion
\mathbb{R}
%EndExpansion
^{n}.$ In order to get $\dot{v}\left(  x\right)  \leq0,$ a direct way of
designing $u$ is proposed as follows%
\begin{equation}
u=-Q\left(  x\right)  f\left(  x\right)  ^{T}\nabla v\left(  x\right)
\label{controller}%
\end{equation}
where $Q:%
%TCIMACRO{\U{211d} }%
%BeginExpansion
\mathbb{R}
%EndExpansion
^{n}\rightarrow%
%TCIMACRO{\U{211d} }%
%BeginExpansion
\mathbb{R}
%EndExpansion
^{l\times l}$ and $Q\left(  x\right)  \geq\epsilon I_{l}>0,$ $\epsilon>0,$
$\forall x\in%
%TCIMACRO{\U{211d} }%
%BeginExpansion
\mathbb{R}
%EndExpansion
^{n}$. Then (\ref{dV}) becomes%
\begin{equation}
\dot{v}\left(  x\right)  =-\nabla v\left(  x\right)  ^{T}f\left(  x\right)
Q\left(  x\right)  f\left(  x\right)  ^{T}\nabla v\left(  x\right)
\leq0.\label{dV1}%
\end{equation}
Substituting (\ref{controller}) into the continuous-time dynamical system
(\ref{dynamical system}) results in%
\begin{equation}
\dot{x}=-f\left(  x\right)  Q\left(  x\right)  f\left(  x\right)  ^{T}\nabla
v\left(  x\right)  \label{dynamics1}%
\end{equation}
with solutions which always satisfy the constraint $c\left(  x\right)  =0.$
The closed-loop system corresponding to the continuous-time dynamical system
(\ref{dynamical system}) and the controller (\ref{controller}) is depicted in
Fig.3.\begin{figure}[h]
\begin{center}
\includegraphics[
scale= 0.8 ]{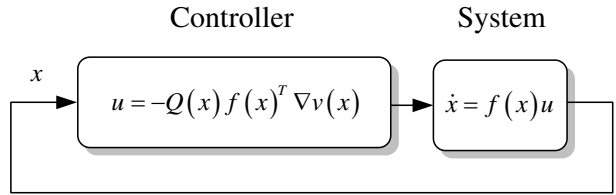}
\end{center}
\caption{Closed-loop control system}%
\end{figure}

\subsection{Convergence Analysis}

Unlike a Lyapunov function, the objective function $v\left(  x\right)  $ is
not required to be positive definite. As a consequence, the conclusions for
Lyapunov functions are not applicable. Instead, the invariance principle is
applied to analyze the behavior of the solution of (\ref{dynamics1}).

\textbf{Theorem 4}. Under\textit{ Assumption 1}, given $x_{0}\in\mathcal{F}$,
if the set $\mathcal{K}$ $=$ $\{x\in%
%TCIMACRO{\U{211d} }%
%BeginExpansion
\mathbb{R}
%EndExpansion
^{n}|v\left(  x\right)  \leq v\left(  x_{0}\right)  ,c\left(  x\right)  =0\}$
is bounded, then the solution of (\ref{dynamics1}) starting at $x_{0}$
approaches $x_{l}^{\ast}\in\mathcal{S}$, where $\mathcal{S}$ $=$
$\{x\in\mathcal{K}|\nabla v\left(  x\right)  ^{T}f\left(  x\right)  =0\}.$ If
in addition $\mathcal{V}_{1}\left(  x_{l}^{\ast}\right)  =\mathcal{V}%
_{2}\left(  x_{l}^{\ast}\right)  ,$ then there must exist a $\lambda^{\ast}$
$=$ $[\lambda_{1}^{\ast}\ \lambda_{2}^{\ast}$ $\cdots$ $\lambda_{m}^{\ast}$
$]^{T}$ $\in%
%TCIMACRO{\U{211d} }%
%BeginExpansion
\mathbb{R}
%EndExpansion
^{m}$ such that $\nabla v\left(  x_{l}^{\ast}\right)  =\sum_{i=1}^{m}%
\lambda_{i}^{\ast}\nabla c_{i}\left(  x_{l}^{\ast}\right)  \ $and $c\left(
x_{l}^{\ast}\right)  =0,$ namely $x_{l}^{\ast}$ is a Karush--Kuhn--Tucker
(KKT) point. Furthermore, if $z^{T}\nabla_{xx}L\left(  x_{l}^{\ast}%
,\lambda^{\ast}\right)  z>0$, for all $z\in\mathcal{V}_{1}\left(  x_{l}^{\ast
}\right)  ,z\neq0,$ then $x_{l}^{\ast}$ is a strict local minimum, where
$L\left(  x,\lambda\right)  =v\left(  x\right)  -\sum_{i=1}^{m}\lambda
_{i}c_{i}\left(  x\right)  .$

\textit{Proof}. The proof is composed of three propositions:
\textit{Proposition 1} is to show that $\mathcal{K}$ is compact and positively
invariant with respect to (\ref{dynamics1}); \textit{Proposition 2} is to show
that the solution of (\ref{dynamics1}) starting at $x_{0}$ approaches
$x_{l}^{\ast}\in\mathcal{S}$; \textit{Proposition 3} is to show that
$x_{l}^{\ast}\in\mathcal{S}$ is a KKT point, further a strict local minimum.
The three propositions are\ proven in \textit{Appendix E}. $\square$

\textbf{Corollary 1}. Suppose that $f\left(  x\right)  $ is chosen as
(\ref{fx}) for $c:%
%TCIMACRO{\U{211d} }%
%BeginExpansion
\mathbb{R}
%EndExpansion
^{n}\rightarrow%
%TCIMACRO{\U{211d} }%
%BeginExpansion
\mathbb{R}
%EndExpansion
^{m}$ and the set $\mathcal{K}$ $=$ $\{x\in%
%TCIMACRO{\U{211d} }%
%BeginExpansion
\mathbb{R}
%EndExpansion
^{n}|v\left(  x\right)  \leq v\left(  x_{0}\right)  ,c\left(  x\right)  =0\}$
is bounded for given $x_{0}\in\mathcal{F}$. Then the solution of
(\ref{dynamics1}) starting at $x_{0}$ approaches $x_{l}^{\ast}\in\mathcal{S}$,
where $\mathcal{S}$ $=$ $\{x\in\mathcal{K}|\nabla v\left(  x\right)
^{T}f\left(  x\right)  =0\},$ where$\ x_{l}^{\ast}$ is a KKT point. In
addition, if $z^{T}\nabla_{xx}L\left(  x_{l}^{\ast},\lambda^{\ast}\right)
z>0$, for all $z\in\mathcal{V}_{1}\left(  x_{l}^{\ast}\right)  ,z\neq0,$ then
$x_{l}^{\ast}$ is a strict local minimum, where $L\left(  x,\lambda\right)
=v\left(  x\right)  -\sum_{i=1}^{m}\lambda_{i}c_{i}\left(  x\right)  .$

\textit{Proof.} Since $\mathcal{V}_{1}\left(  x_{l}^{\ast}\right)
=\mathcal{V}_{2}\left(  x_{l}^{\ast}\right)  $ by \textit{Theorem 3},\textit{
}the remainder of the proof is the same as that of\textit{ Theorem 4}.\textbf{
}$\square$

\textbf{Corollary 2}. Consider the following equality-constrained optimization
problem%
\begin{equation}
\underset{x\in%
%TCIMACRO{\U{211d} }%
%BeginExpansion
\mathbb{R}
%EndExpansion
^{n}}{\min}v\left(  x\right)  ,\text{ s.t. }Ax=b. \label{convex}%
\end{equation}
If (i) $v\left(  x\right)  $ is convex and twice continuously differentiable,
(ii) $A\in%
%TCIMACRO{\U{211d} }%
%BeginExpansion
\mathbb{R}
%EndExpansion
^{p\times n}$ with rank$A<n,$ (iii) $\mathcal{K=}\{x\in%
%TCIMACRO{\U{211d} }%
%BeginExpansion
\mathbb{R}
%EndExpansion
^{n}|v\left(  x\right)  \leq v\left(  x_{0}\right)  ,Ax=b\}$ is bounded, then
the solution of (\ref{dynamics1}) with $f\left(  x\right)  =A^{\bot}$ starting
at any $x_{0}\in\mathcal{F}$ approaches $x^{\ast}$.

\textit{Proof.} The solution of (\ref{dynamics1}) starting at $x_{0}$
approaches $x_{l}^{\ast}\in\mathcal{S}.$ Since rank$A<n,$ we have
$\mathcal{V}_{1}\left(  x_{l}^{\ast}\right)  =\mathcal{V}_{2}\left(
x_{l}^{\ast}\right)  \neq\varnothing.$ Since the equality constrained
optimization problem (\ref{convex}) is convex, a KKT point $x_{l}^{\ast}$\ is
a global minimum $x^{\ast}$ of the problem (\ref{convex}). The remainder of
proof is the same as that of\textit{ Theorem 4}.\textbf{ }$\square$

\textbf{Remark 6}. If $\mathcal{K}$ is not a bounded set, then $\mathcal{S}$
defined in \textit{Theorem 4} may be empty. Therefore, the boundedness of the
set $\mathcal{K}$ is necessary. For example, $v\left(  x\right)  =x_{1}%
+x_{2},$ s.t. $c\left(  x\right)  =x_{1}-x_{2}=0$. The set $\mathcal{K}%
=\left\{  \left.  x\in%
%TCIMACRO{\U{211d} }%
%BeginExpansion
\mathbb{R}
%EndExpansion
^{2}\right\vert x_{1}+x_{2}\leq v\left(  x_{0}\right)  ,x_{1}-x_{2}=0\right\}
$ is unbounded. According to \textit{Theorem 1}, we have $f\left(  x\right)
=[1$ $1]^{T}.$ In this case, $\nabla v\left(  x\right)  ^{T}f\left(  x\right)
=2\neq0\ $and then the set $\mathcal{S}$ is empty.

\subsection{A Modified Closed-Loop Dynamical System}

Although the proposed approach ensures that the solutions satisfy the
constraint, this approach may fail if $x_{0}\notin\mathcal{F}$ or if numerical
algorithms are used to compute the solutions. Moreover, if the impulse
function $\delta$ is approximated, then the constraints will also be violated.
With these results, the following modified closed-loop dynamical system is
proposed to amend this situation.

Similar to \cite{Yamashita(1980)}, we introduce the term $-\rho\nabla c\left(
x\right)  c\left(  x\right)  $ into (\ref{dynamics1}), resulting in%
\begin{equation}
\dot{x}=-\rho\nabla c\left(  x\right)  c\left(  x\right)  -f\left(  x\right)
Q\left(  x\right)  f\left(  x\right)  ^{T}\nabla v\left(  x\right)  ,x\left(
0\right)  =x_{0} \label{modifieddynamics1}%
\end{equation}
where $\rho>0$. Define $v_{c}\left(  x\right)  =c\left(  x\right)
^{T}c\left(  x\right)  .$ Then
\[
\dot{v}_{c}\left(  x\right)  =-\rho c\left(  x\right)  ^{T}\nabla c\left(
x\right)  ^{T}\nabla c\left(  x\right)  c\left(  x\right)  \leq0,
\]
where $\nabla c\left(  x\right)  ^{T}f\left(  x\right)  \equiv0$ is utilized.
If the impulse function $\delta$ is approximated, then $\nabla c\left(
x\right)  ^{T}f\left(  x\right)  \approx0$ and can be ignored in practice.
Therefore, the solutions of (\ref{modifieddynamics1}) will tend to the
feasible set $\mathcal{F}$ if $\nabla c\left(  x\right)  $ is of full column
rank. Once $c\left(  x\right)  =0,$ the modified dynamical system
(\ref{modifieddynamics1}) degenerates to (\ref{dynamics1}). The
self-correcting feature enables the step size to be automatically controlled
in the numerical integration process or to tolerate uncertainties when the
differential equation is realized by using analog hardware.

\textbf{Remark 7}. The matrix $Q\left(  x\right)  $ plays a role in
coordinating the convergence rate of all states by minimizing the condition
number of the matrix functions like $f\left(  x\right)  Q\left(  x\right)
f\left(  x\right)  ^{T}$. Moreover, it also plays a role in avoiding
instability in the numerical solution of differential equations by normalizing
the Lipschitz condition of functions like $f\left(  x\right)  Q\left(
x\right)  f\left(  x\right)  ^{T}\nabla v\left(  x\right)  .$ Concrete
examples are given in the following section.

\section{Illustrative Examples}

\subsection{Estimate of Attraction Domain}

For a given Lyapunov function, the crucial step in any procedure for
estimating the attraction domain is determining the optimal estimate. Consider
the system of differential\ equations:%
\begin{equation}
\dot{x}=Ax+g\left(  x\right)  \label{marginsys}%
\end{equation}
where $x\in%
%TCIMACRO{\U{211d} }%
%BeginExpansion
\mathbb{R}
%EndExpansion
^{n}$ is the state vector, $A\in%
%TCIMACRO{\U{211d} }%
%BeginExpansion
\mathbb{R}
%EndExpansion
^{n\times n}$ is a Hurwitz matrix, and $g:%
%TCIMACRO{\U{211d} }%
%BeginExpansion
\mathbb{R}
%EndExpansion
^{n}\rightarrow%
%TCIMACRO{\U{211d} }%
%BeginExpansion
\mathbb{R}
%EndExpansion
^{n}$ is a vector function. Let $v\left(  x\right)  =x^{T}Px$ be a given
quadratic Lyapunov function for the origin of (\ref{marginsys}), i.e., $P\in%
%TCIMACRO{\U{211d} }%
%BeginExpansion
\mathbb{R}
%EndExpansion
^{n\times n}$ is a positive-definite matrix such that $A^{T}P+PA<0$. Then the
largest ellipsoidal estimate of the attraction domain of the origin can be
computed via the following equality-constrained optimization problem
\cite{Graziano(2003)}:%
\begin{equation}
\underset{x\in%
%TCIMACRO{\U{211d} }%
%BeginExpansion
\mathbb{R}
%EndExpansion
^{n}\backslash\left\{  0\right\}  }{\min}x^{T}Px\text{ s.t. }x^{T}P\left[
Ax+g\left(  x\right)  \right]  =0.\nonumber
\end{equation}

Since $\{x\in%
%TCIMACRO{\U{211d} }%
%BeginExpansion
\mathbb{R}
%EndExpansion
^{n}|x^{T}Px\leq x_{0}^{T}Px_{0}\}$ is bounded, the subset
\[
\mathcal{K}=\{x\in%
%TCIMACRO{\U{211d} }%
%BeginExpansion
\mathbb{R}
%EndExpansion
^{n}|x^{T}Px\leq x_{0}^{T}Px_{0},x^{T}P\left[  Ax+g\left(  x\right)  \right]
=0\}
\]
is bounded no matter what $g$ is.

For simplicity, consider (\ref{marginsys}) with $x=[x_{1}$ $x_{2}]^{T}\in%
%TCIMACRO{\U{211d} }%
%BeginExpansion
\mathbb{R}
%EndExpansion
^{2},$ $A=-I_{2},$ $P=I_{2}$ and $g\left(  x\right)  =\left(  \sigma\left(
x\right)  +1\right)  [x_{1}$ $x_{2}]^{T},$ where $\sigma\left(  x\right)
=\left(  x_{1}+x_{2}+2\right)  \left(  \left(  x_{2}+1\right)  -0.1\left(
x_{1}+1\right)  ^{2}\right)  .$ Then the optimization problem is formulated as%
\begin{equation}
\underset{x\in%
%TCIMACRO{\U{211d} }%
%BeginExpansion
\mathbb{R}
%EndExpansion
^{2}\backslash\left\{  0\right\}  }{\min}x_{1}^{2}+x_{2}^{2}\text{ s.t.
}\left(  x_{1}^{2}+x_{2}^{2}\right)  \sigma\left(  x\right)  =0.\nonumber
\end{equation}
Since $x\neq0,$ the problem is further formulated as
\begin{equation}
\underset{x\in%
%TCIMACRO{\U{211d} }%
%BeginExpansion
\mathbb{R}
%EndExpansion
^{2}}{\min}v\left(  x\right)  =x_{1}^{2}+x_{2}^{2}\text{ s.t. }\sigma\left(
x\right)  =0.\nonumber
\end{equation}
Then
\begin{align*}
\nabla v\left(  x\right)   &  =[2x_{1}2x_{2}]^{T}\\
\nabla c\left(  x\right)   &  =\left[
\begin{array}
[c]{c}%
d_{2}-0.1d_{1}^{2}-0.2d_{1}d_{3}\\
d_{2}-0.1d_{1}^{2}+d_{3}%
\end{array}
\right] \\
d_{1}  &  =x_{1}+1,d_{2}=x_{2}+1,d_{3}=x_{1}+x_{2}+2.
\end{align*}
In this example, we adopt the modified dynamics (\ref{modifieddynamics1}),
where $f$ is chosen as (\ref{fx}) with$\ \delta\left(  x\right)
=e^{-\gamma\left\vert x\right\vert },$ \ and the parameters$\ $are chosen
as$\ \gamma=10,\rho=Q=20\left/  \left\Vert \nabla cc-ff^{T}\nabla v\right\Vert
\right.  .$ We solve the differential equation (\ref{modifieddynamics1}) by
using the MATLAB function \textquotedblleft ode45\textquotedblright\ with
\textquotedblleft variable-step\footnote{In this section, all computation is
performed by MATLAB 6.5 on a personal computer (Asus x8ai) with Intel core Duo
2 Processor at 2.2GHz.}\textquotedblright. Compared with the MATLAB optimal
constrained nonlinear multivariate function \textquotedblleft
fmincon\textquotedblright, we derive the comparisons in Table
1.\begin{figure}[ptbh]
\centering
\begin{gather*}
\text{{\small TABLE 1. COMPUTED RESULT FOR EXAMPLE 1}}\\
\text{%
\begin{tabular}
[c]{ccccc}\hline\hline
{\small Method} & {\small Initial Point} & {\small Solution} & {\small Optimal
Value} & {\small cpu time (sec.)}\\\hline
{\small Matlab fmincon} & {\small [-3 1]}$^{\text{{\scriptsize T}}}$ &
{\small [-1 -1]}$^{\text{{\scriptsize T}}}$ & {\small 2.0000} & {\small Not
Available}\\
{\small New method} & {\small [-3 1]}$^{\text{{\scriptsize T}}}$ &
{\small [0.2062 -0.8546]}$^{\text{{\scriptsize T}}}$ & {\small 0.7729} &
{\small 0.125}\\\hline
{\small Matlab fmincon} & {\small [2 -4]}$^{\text{{\scriptsize T}}}$ &
{\small [-1 -1]}$^{\text{{\scriptsize T}}}$ & {\small 2.0000} & {\small Not
Available}\\
{\small New method} & {\small [2 -4]}$^{\text{{\scriptsize T}}}$ &
{\small [0.2062 -0.8545]}$^{\text{{\scriptsize T}}}$ & {\small 0.7726} &
{\small 0.0940}\\\hline
{\small Matlab fmincon} & {\small [1 -4]}$^{\text{{\scriptsize T}}}$ &
{\small [0.2143 -0.8533]}$^{\text{{\scriptsize T}}}$ & {\small 0.7740} &
{\small 0.2030}\\
{\small New method} & {\small [1 -4]}$^{\text{{\scriptsize T}}}$ &
{\small [0.2056 -0.8550]}$^{\text{{\scriptsize T}}}$ & {\small 0.7733} &
{\small 0.1100}\\\hline\hline
\end{tabular}
.}%
\end{gather*}
\end{figure}

The point $x_{s}=[-1$ $-1]^{T}$ is a singular point, at which $\nabla c\left(
x_{s}\right)  =[0$ $0]^{T}.$ As shown in Table 1, under initial points $[-3$
$1]^{T}\in\mathcal{F}$ and $[2$ $-4]^{T}\in\mathcal{F},$ the MATLAB function
fails to find the minimum and stops at the singular point, whereas the
proposed approach still finds the minimum. Under initial point $[1$
$-4]^{T}\notin\mathcal{F},$ the proposed approach can still find the minimum,
similar to the MATLAB function. Under a different initial value, the
evolutions of (\ref{modifieddynamics1}) are shown in Fig.4. As shown, once
close to the singular point $[-1$ $-1]^{T}$, the solutions of
(\ref{modifieddynamics1}) change direction and then move to the minimum
$x_{l}^{\ast}=[0.2061\ -0.8545]^{T}$. Compared with the discrete optimal
methods offered by MATLAB, these results show that the proposed approach
avoids convergence to a singular point. Moreover, the proposed approach is
comparable with currently available conventional discrete optimal methods and
facilitates even faster convergence. The latter conclusion is consistent with
that proposed in \cite{Tanabe(1980)},\cite{Brown(1989)}. \begin{figure}[ptbh]
%Requires \usepackage{graphicx}
\centering
\includegraphics[width=0.8\linewidth]{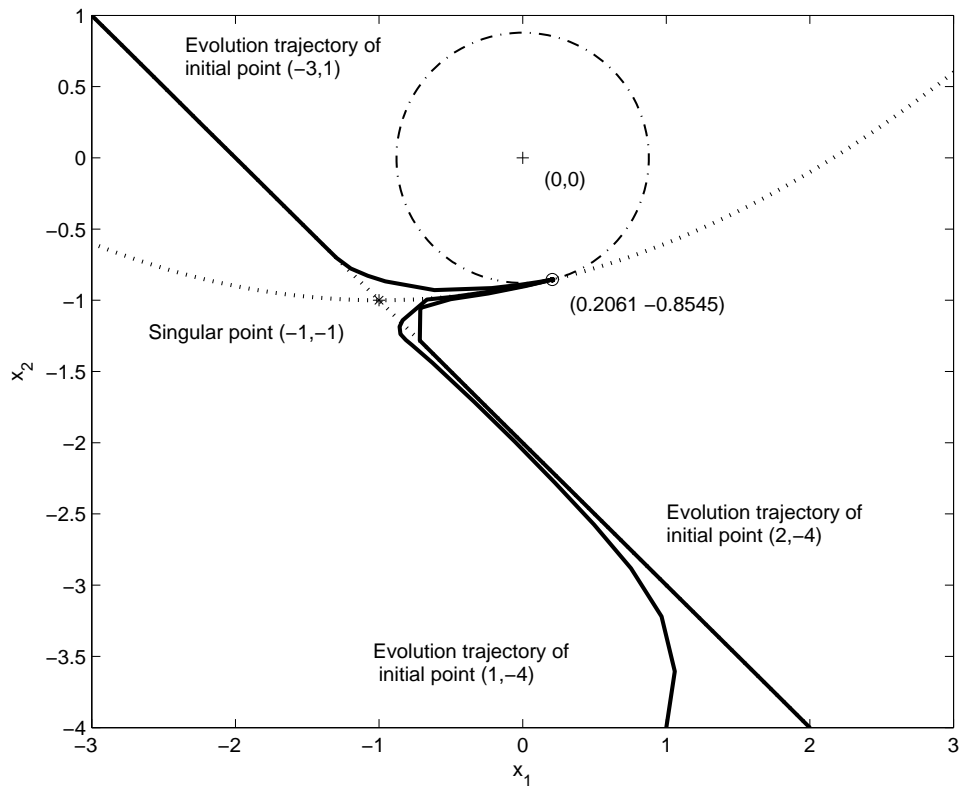}\caption{Optimization for
estimate of attraction domain. Solution Evolution (solid line), Constraint
(dot line), Objective (dash-dot line).}%
\end{figure}

\subsection{Estimate of Essential Matrix}

For simplicity, assume that images are taken by two identical pin-hole cameras
with focal length equal to one. The two cameras are specified by the camera
centers $C_{1},C_{2}\in%
%TCIMACRO{\U{211d} }%
%BeginExpansion
\mathbb{R}
%EndExpansion
^{3}$ and attached orthogonal camera frames $\left\{  e_{1},e_{2}%
,e_{3}\right\}  $ and $\left\{  e_{1}^{\prime},e_{2}^{\prime},e_{3}^{\prime
}\right\}  $, respectively. Denote $T=C_{2}-C_{1}\in%
%TCIMACRO{\U{211d} }%
%BeginExpansion
\mathbb{R}
%EndExpansion
^{3}$ to be the translation from the first camera to the second and $R\in%
%TCIMACRO{\U{211d} }%
%BeginExpansion
\mathbb{R}
%EndExpansion
^{3\times3}$ to be the rotation matrix from the basis vectors $\left\{
e_{1},e_{2},e_{3}\right\}  $ to $\left\{  e_{1}^{\prime},e_{2}^{\prime}%
,e_{3}^{\prime}\right\}  $, expressed with respect to the basis $\left\{
e_{1},e_{2},e_{3}\right\}  .$ Then, it is well known in the computer vision
literature \cite{Hartley(2003)} that two corresponding image points are
represented as follows:%
\begin{align}
m_{1,k} &  =\frac{1}{M_{k}\left(  3\right)  }M_{k},\nonumber\\
m_{2,k} &  =\frac{1}{M_{k}^{\prime}\left(  3\right)  }M_{k}^{\prime
},k=1,2,\cdots,N\label{image points}%
\end{align}
where $M_{k},M_{k}^{\prime}$ represent the positions of the $k$th point
expressed in the two camera frames $\left\{  e_{1},e_{2},e_{3}\right\}  $ to
$\left\{  e_{1}^{\prime},e_{2}^{\prime},e_{3}^{\prime}\right\}  ,$
respectively; $M_{k}\left(  3\right)  ,M_{k}^{\prime}\left(  3\right)  $
represent the third element of vectors $M_{k},M_{k}^{\prime},$ respectively.
They have the relationship $M_{k}=RM_{k}^{\prime}+T,$ $k=1,2,\cdots,N.$ These
corresponding image points satisfy the socalled epipolar constraint \cite[p.
257]{Hartley(2003)}:%
\begin{equation}
m_{1,k}^{T}Em_{2,k}=0,k=1,2,\cdots,N\label{Essential}%
\end{equation}
where $E=\left[  T\right]  _{\times}R\ $is known as the \emph{essential
matrix}.\begin{figure}[h]
\begin{center}
\includegraphics[
scale=0.9 ]{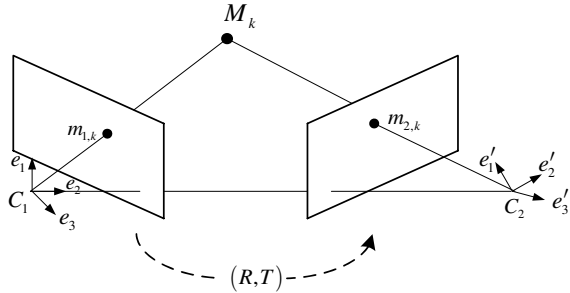}
\end{center}
\caption{Epipolar geometry}%
\end{figure}

By using the direct product $\otimes$ and the $\operatorname{vec}\left(
\cdot\right)  $\ operation, the equations in (\ref{Essential}) are equivalent
to%
\begin{equation}
A\varphi=0_{N\times1} \label{EssentialVec}%
\end{equation}
where%
\begin{align}
A  &  =\left[
\begin{array}
[c]{c}%
m_{2,1}^{T}\otimes m_{1,1}^{T}\\
\vdots\\
m_{2,N}^{T}\otimes m_{1,N}^{T}%
\end{array}
\right]  \in%
%TCIMACRO{\U{211d} }%
%BeginExpansion
\mathbb{R}
%EndExpansion
^{N\times9},\nonumber\\
\varphi &  =\text{vec}\left(  \left[  T\right]  _{\times}R\right)  . \label{A}%
\end{align}
In practice, these image points $m_{1,k}$ and $m_{2,k}$ are subject to noise,
$k=1,2,\cdots,N$. Therefore, $T$ and $R$ are often solved by the following
optimization problem%
\begin{gather}
\underset{x\in%
%TCIMACRO{\U{211d} }%
%BeginExpansion
\mathbb{R}
%EndExpansion
^{12}}{\min}v\left(  x\right)  =\frac{1}{2}\varphi\left(  x\right)  ^{T}%
A^{T}A\varphi\left(  x\right) \nonumber\\
\text{s.t. }\frac{1}{2}\left(  \left\Vert T\right\Vert ^{2}-1\right)
=0\nonumber\\
\text{ \ \ \ \ }\frac{1}{2}\left(  R^{T}R-I_{3}\right)  =0_{3\times3}
\label{essential matrix optimal}%
\end{gather}
where $x=[T^{T}$ vec$^{T}\left(  R\right)  ]^{T}\in%
%TCIMACRO{\U{211d} }%
%BeginExpansion
\mathbb{R}
%EndExpansion
^{12}$. This is an equality-constrained optimization considered here. In the
following, the proposed approach is applied to the optimization problem
(\ref{essential matrix optimal}). By\textit{ Theorem 2}, the projection matrix
for the constraint $\frac{1}{2}\left(  \left\Vert T\right\Vert ^{2}-1\right)
=0$ is%
\[
f=I_{3}-\frac{TT^{T}}{\delta\left(  \left\Vert T\right\Vert ^{2}\right)
+\left\Vert T\right\Vert ^{2}}.
\]
Since $\left\Vert T\right\Vert ^{2}=1$ has to be satisfied exactly or
approximately, then $\delta\left(  \left\Vert T\right\Vert ^{2}\right)  =0.$
So, the projection matrix for the constraint is
\[
f=I_{3}-T^{T}\left/  \left\Vert T\right\Vert ^{2}\right.  .
\]
Then the constraint is transformed into
\[
\dot{T}=\left(  I_{3}-TT^{T}\left/  \left\Vert T\right\Vert ^{2}\right.
\right)  u_{1},
\]
where$\ u_{1}\in%
%TCIMACRO{\U{211d} }%
%BeginExpansion
\mathbb{R}
%EndExpansion
^{3}$. By (\ref{rotation}), the constraint $\frac{1}{2}\left(  R^{T}%
R-I_{3}\right)  =0_{3\times3}\ $is transformed into
\[
\dot{R}=\left[  u_{2}\right]  _{\times}R,
\]
where $u_{2}\in%
%TCIMACRO{\U{211d} }%
%BeginExpansion
\mathbb{R}
%EndExpansion
^{3}.$ Furthermore, the equation above is rewritten as
\[
\text{vec}\left(  \dot{R}\right)  =\left(  R^{T}\otimes I_{3}\right)  Hu_{2}.
\]
Then the continuous-time dynamical system, whose solutions always satisfy the
equality constraints $\frac{1}{2}\left(  \left\Vert T\right\Vert
^{2}-1\right)  $ $=$ $0\ $and $\frac{1}{2}\left(  R^{T}R-I_{3}\right)  $ $=$
$0_{3\times3}$, is expressed as (\ref{dynamical system}) with%
\begin{align}
f\left(  x\right)   &  =\left[
\begin{array}
[c]{cc}%
I_{3}-TT^{T}\left/  \left\Vert T\right\Vert ^{2}\right.  & 0_{3\times3}\\
0_{9\times3} & \left(  R^{T}\otimes I_{3}\right)  H
\end{array}
\right]  \in%
%TCIMACRO{\U{211d} }%
%BeginExpansion
\mathbb{R}
%EndExpansion
^{12\times6},\nonumber\\
u  &  =\left[
\begin{array}
[c]{c}%
u_{1}\\
u_{2}%
\end{array}
\right]  \in%
%TCIMACRO{\U{211d} }%
%BeginExpansion
\mathbb{R}
%EndExpansion
^{6}. \label{f}%
\end{align}
If the initial value $\left\Vert T\left(  0\right)  \right\Vert ^{2}=1$ and
$R\left(  0\right)  ^{T}R\left(  0\right)  =I_{3},$ then all solutions of
(\ref{dynamical system}) satisfy the equality constraints. Since $\nabla
v\left(  x\right)  =[%
\begin{array}
[c]{cc}%
\left(  R^{T}\otimes I_{3}\right)  H & I_{3}\otimes\left[  T\right]  _{\times}%
\end{array}
]^{T}A^{T}A\varphi,$ the time derivative of $v\left(  x\right)  $ along the
solutions of (\ref{dynamical system}) is
\[
\dot{v}\left(  x\right)  =-\varphi^{T}A^{T}A\Theta\left(  x\right)
^{T}Q\left(  x\right)  \Theta\left(  x\right)  A^{T}A\varphi\leq0,
\]
where
\[
\Theta\left(  x\right)  =\left[
\begin{array}
[c]{c}%
\left(  I_{3}-TT^{T}\left/  \left\Vert T\right\Vert ^{2}\right.  \right)
^{T}H^{T}\left(  R^{T}\otimes I_{3}\right)  ^{T}\\
H^{T}\left(  R^{T}\otimes I_{3}\right)  ^{T}\left(  I_{3}\otimes\left[
T\right]  _{\times}\right)  ^{T}%
\end{array}
\right]  \in%
%TCIMACRO{\U{211d} }%
%BeginExpansion
\mathbb{R}
%EndExpansion
^{6\times9}.
\]

The simplest way of choosing $Q\left(  x\right)  $ is $Q\left(  x\right)
\equiv I_{6}$. In this case, the eigenvalues of the matrix $A\Theta^{T}\left(
x\right)  \Theta\left(  x\right)  A^{T}$ are often ill-conditioned, namely
\[
\lambda_{\text{min}}\left(  A\Theta^{T}\left(  x\right)  \Theta\left(
x\right)  A^{T}\right)  \ll\lambda_{\text{max}}\left(  A\Theta^{T}\left(
x\right)  \Theta\left(  x\right)  A^{T}\right)  .
\]
Convergence rates of the components of $A\varphi\left(  x\right)  $ depend on
the eigenvalues of $A\Theta^{T}\left(  x\right)  Q\left(  x\right)
\Theta\left(  x\right)  A^{T}.$ As a consequence, some components of
$A\varphi$ converge fast, while the other may converge slowly. This leads to
poor asymptotic performance of the closed-loop system. It is expected that
each component of $A\varphi$ can converge at the same speed as far as
possible. Suppose that there exists a $\bar{Q}\left(  x\right)  $ such that
\[
A\Theta^{T}\left(  x\right)  \bar{Q}\left(  x\right)  \Theta\left(  x\right)
A^{T}=I_{9}.
\]
Then
\[
\dot{v}\left(  x\right)  \leq-\varphi^{T}A^{T}A\varphi\leq0.
\]
By \textit{Theorem 4}, $x\ $will approach the set $\left\{  \left.  x\in%
%TCIMACRO{\U{211d} }%
%BeginExpansion
\mathbb{R}
%EndExpansion
^{n}\right\vert A\varphi\left(  x\right)  =0\right\}  ,$ each element of which
is a global minimum since $v\left(  x\right)  =0$ in the set. Moreover, each
component of $A\varphi$ converges at a similar speed. However, it is difficult
to obtain such a $\bar{Q}\left(  x\right)  $, since the number of degrees of
freedom of $\bar{Q}\left(  x\right)  \in%
%TCIMACRO{\U{211d} }%
%BeginExpansion
\mathbb{R}
%EndExpansion
^{6\times6}$ is less than the number of elements of $I_{9}$. A modified way is
to make $A\Theta^{T}\left(  x\right)  Q\left(  x\right)  \Theta\left(
x\right)  A^{T}\approx I_{9}.$ A natural choice is proposed as follows%
\begin{equation}
Q\left(  x\right)  =\mu\left(  \left(  \Theta\left(  x\right)  A^{T}%
A\Theta\left(  x\right)  ^{T}\right)  ^{\dag}+\epsilon I_{6}\right)  \label{Q}%
\end{equation}
where $\mu>0,$ $\left(  \Theta\left(  x\right)  A^{T}A\Theta^{T}\left(
x\right)  \right)  ^{\dag}$ denotes the Moore Penrose inverse of
$\Theta\left(  x\right)  A^{T}A\Theta^{T}\left(  x\right)  $. The matrix
$\epsilon I_{6}$ is to make $Q\left(  x\right)  $ positive definite, where
$\epsilon$ is a small positive real. From the procedure above, $\left(
\Theta\left(  x\right)  A^{T}A\Theta^{T}\left(  x\right)  \right)  ^{\dag}$
needs to be computed every time. This however will cost much time. A
time-saving way is to update $Q\left(  x\right)  $ at a reasonable interval.
Then (\ref{dynamics1}) becomes%
\begin{equation}
\dot{x}=-\mu f\left(  x\right)  \left(  \left(  \Theta\left(  x\right)
A^{T}A\Theta\left(  x\right)  ^{T}\right)  ^{\dag}+\epsilon I_{6}\right)
\Theta\left(  x\right)  A^{T}A\varphi\left(  x\right)  \label{dynamics2}%
\end{equation}
where $f\left(  x\right)  $ is defined in (\ref{f}). The differential equation
can be solved by Runge-Kutta methods, etc. The solutions of (\ref{dynamics2})
satisfy the constraints, where $x=[T^{T}$ vec$\left(  R\right)  ^{T}]^{T}.$
Moreover, the dynamic system will reach some final resting state eventually.

Suppose that there exist 6 points in the field of view, whose positions are
expressed in the first camera frame as follows: $M_{1}=[-1$ $1$ $1]^{T},$
$M_{2}=[2$ $0$ $1]^{T},$ $M_{3}=[1$ $-1$ $1]^{T},$ $M_{4}=[-1$ $-1$ $1]^{T},$
$M_{5}=[1$ $1$ $1]^{T},$ $M_{6}=[-1$ $3$ $1]^{T}.$ Compared with the first
camera frame, the second camera frame has translated and rotated with%
\[
\bar{T}=\left[
\begin{array}
[c]{c}%
1\\
1\\
-1
\end{array}
\right]  ,\bar{R}=\left[
\begin{array}
[c]{ccc}%
0.9900 & -0.0894 & 0.1088\\
0.0993 & 0.9910 & -0.0894\\
-0.0998 & 0.0993 & 0.9900
\end{array}
\right]  .
\]

The image points are generated by (\ref{image points}). Using the generated
image points, we obtain $A$ by (\ref{A}). Setting the initial value as follows
$T\left(  0\right)  =[0$ $0$ $1]^{T},$ $R\left(  0\right)  =I_{3},$ $\mu=20,$
$\epsilon=0.01.$ We solve the differential equation (\ref{modifieddynamics1})
by using MATLAB function \textquotedblleft ode45\textquotedblright\ with
\textquotedblleft variable-step\textquotedblright. Compared with MATLAB
optimal constrained nonlinear multivariate function \textquotedblleft
fmincon\textquotedblright, we have the following comparisons:%
\begin{gather*}
\text{{\small TABLE 2. COMPUTED RESULT FOR EXAMPLE 2}}\\
\text{%
\begin{tabular}
[c]{ccc}\hline\hline
{\small Method} & $\left\Vert R^{\ast T}\bar{R}-I_{3}\right\Vert $ &
{\small cpu time (sec.)}\\\hline
{\small MATLAB fmincon} & {\small 1.2469e-004} & {\small 0.2500}\\
{\small New Approach} & {\small 1.8784e-005} & {\small 0.1400}\\\hline\hline
\end{tabular}
.}%
\end{gather*}
As shown in Table 2, the proposed approach requires less time to achieve a
higher accuracy. Given that $v\left(  x^{\ast}\right)  =0$, the solution is a
global minimum. The evolution of each element of $x$ is shown in Fig.5. The
state eventually reaches a rest state at a similar speed. With different
initial values, several other simulations are also implemented. Based on the
results, the proposed algorithm has met the expectations.\begin{figure}[h]
\begin{center}
\includegraphics[
scale=0.5]{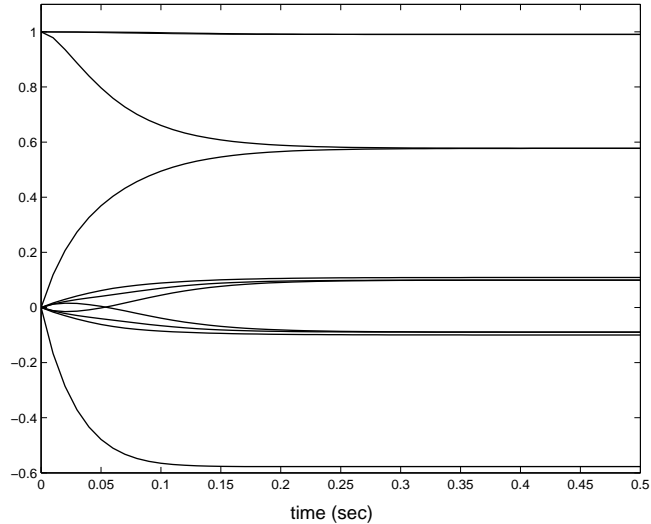}
\end{center}
\caption{Evolvement of the state}%
\end{figure}\ 

\section{Conclusions}

An approach to continuous-time, equality-constrained optimization based on a
new projection matrix is proposed for the determination of local minima. With
the transformation of the equality constraint into a continuous-time dynamical
system, the class of equality-constrained optimization is formulated as a
control problem. The resultant approach is more general than the existing
control theoretic approaches. Thus, the proposed approach serves as a
potential bridge between the optimization and control theories. Compared with
other standard discrete-time methods, the proposed approach avoids convergence
to a singular point and facilitates faster convergence through numerical
integration on a digital computer.

\section*{Appendix}

\emph{A. Kronecker Product and Vec}

The symbol vec$(X)$ is the column vector obtained by stacking the second
column of $X$ under the first, and then the third, and so on. With $X=\left[
x_{ij}\right]  \in%
%TCIMACRO{\U{211d} }%
%BeginExpansion
\mathbb{R}
%EndExpansion
^{n\times m}$, the \textit{Kronecker product }$X\otimes Y$ is the matrix%
\[
X\otimes Y=\left[
\begin{array}
[c]{ccc}%
x_{11}Y & \cdots & x_{1m}Y\\
\vdots & \ddots & \vdots\\
x_{n1}Y & \cdots & x_{nm}Y
\end{array}
\right]  .
\]
In fact, we have the following relationships vec$(XYZ)=\left(  Z^{T}\otimes
X\right)  $vec$(Y)$ \cite[p. 318]{Helmke (1994)}.

\emph{B. Skew-Symmetric Matrix}

The cross product of two vectors $x\in%
%TCIMACRO{\U{211d} }%
%BeginExpansion
\mathbb{R}
%EndExpansion
^{3}$ and $y\in%
%TCIMACRO{\U{211d} }%
%BeginExpansion
\mathbb{R}
%EndExpansion
^{3}$ is denoted by $x\times y=\left[  x\right]  _{\times}y,$ where the symbol
$\left[  \cdot\right]  _{\times}:$ $%
%TCIMACRO{\U{211d} }%
%BeginExpansion
\mathbb{R}
%EndExpansion
^{3}$ $\rightarrow$ $%
%TCIMACRO{\U{211d} }%
%BeginExpansion
\mathbb{R}
%EndExpansion
^{3\times3}$ is defined as \cite[p. 194]{Alberto Isdori (2003)}:%
\[
\left[  x\right]  _{\times}\triangleq\left[
\begin{array}
[c]{ccc}%
0 & -x_{3} & x_{2}\\
x_{3} & 0 & -x_{1}\\
-x_{2} & x_{1} & 0
\end{array}
\right]  \in%
%TCIMACRO{\U{211d} }%
%BeginExpansion
\mathbb{R}
%EndExpansion
^{3\times3}.
\]
By the definition of $\left[  x\right]  _{\times},$ we have$\ x\times
x=\left[  x\right]  _{\times}x=0_{3\times1},$ $\forall x\in%
%TCIMACRO{\U{211d} }%
%BeginExpansion
\mathbb{R}
%EndExpansion
^{3}\ $and%
\begin{align*}
\text{vec}\left(  \left[  x\right]  _{\times}\right)   &  =Hx,\\
H  &  =\left[
\begin{array}
[c]{ccccccccc}%
0 & 0 & 0 & 0 & 0 & 1 & 0 & -1 & 0\\
0 & 0 & -1 & 0 & 0 & 0 & 1 & 0 & 0\\
0 & 1 & 0 & -1 & 0 & 0 & 0 & 0 & 0
\end{array}
\right]  ^{T}.
\end{align*}

\emph{C. Proof of Lemma 1}

Since $\delta\left(  \left\Vert L\right\Vert ^{2}\right)  +\left\Vert
L\right\Vert ^{2}=1$ if $L=0$ and $\delta\left(  \left\Vert L\right\Vert
^{2}\right)  +\left\Vert L\right\Vert ^{2}=\left\Vert L\right\Vert ^{2}$ if
$L\neq0,$ we have $\delta\left(  \left\Vert L\right\Vert ^{2}\right)
+\left\Vert L\right\Vert ^{2}\neq0$, $\forall L\in%
%TCIMACRO{\U{211d} }%
%BeginExpansion
\mathbb{R}
%EndExpansion
^{n}.$ According to this, we have the following relationship%
\begin{align*}
&  L^{T}\left(  I_{n}-LL^{T}\left/  \left(  \delta\left(  \left\Vert
L\right\Vert ^{2}\right)  +\left\Vert L\right\Vert ^{2}\right)  \right.
\right) \\
&  =L^{T}-L^{T}\left\Vert L\right\Vert ^{2}\left/  \left(  \delta\left(
\left\Vert L\right\Vert ^{2}\right)  +\left\Vert L\right\Vert ^{2}\right)
\right. \\
&  \equiv0,\text{ }\forall L\in%
%TCIMACRO{\U{211d} }%
%BeginExpansion
\mathbb{R}
%EndExpansion
^{n}.
\end{align*}
This implies that $L^{T}z=0,$ $\forall z\in\mathcal{W}_{2}$, namely
$\mathcal{W}_{2}\subseteq\mathcal{W}_{1}$. On the other hand, any
$z\in\mathcal{W}_{1}$ is rewritten as
\[
z=\left(  I_{n}-LL^{T}\left/  \left(  \delta\left(  \left\Vert L\right\Vert
^{2}\right)  +\left\Vert L\right\Vert ^{2}\right)  \right.  \right)  z
\]
where $L^{T}z=0$ is utilized. Hence $\mathcal{W}_{1}\subseteq\mathcal{W}_{2}.$
Consequently, $\mathcal{W}_{1}=\mathcal{W}_{2}.$

\emph{D. Proof of Theorem 3}

Denote%
\begin{align*}
\mathcal{V}_{1}^{j}  &  =\{z\in%
%TCIMACRO{\U{211d} }%
%BeginExpansion
\mathbb{R}
%EndExpansion
^{n}|\nabla c_{i}^{T}z=0,i=1,\cdots,j,j\leq m\}\\
\mathcal{V}_{2}^{j}  &  =\{z\in%
%TCIMACRO{\U{211d} }%
%BeginExpansion
\mathbb{R}
%EndExpansion
^{n}|z=f_{j}u_{j},u_{j}\in%
%TCIMACRO{\U{211d} }%
%BeginExpansion
\mathbb{R}
%EndExpansion
^{n},j\leq m\}.
\end{align*}
First, by \textit{Theorem 2},\textit{ }it is easy to see that the conclusions
are satisfied with $j=1$. Assume $\mathcal{V}_{1}^{k-1}$ $=$ $\mathcal{V}%
_{2}^{k-1}$ and then prove that $\mathcal{V}_{1}^{k}$ $=$ $\mathcal{V}_{2}%
^{k}$ holds. If so, then we can conclude this proof. By $\mathcal{V}_{1}%
^{k-1}\left(  x\right)  $ $=$ $\mathcal{V}_{2}^{k-1}\left(  x\right)  ,$ we
have%
\begin{align*}
\mathcal{V}_{1}^{k}  &  =\{z\in%
%TCIMACRO{\U{211d} }%
%BeginExpansion
\mathbb{R}
%EndExpansion
^{n}|\nabla c_{k}^{T}z=0,z\in\mathcal{V}_{1}^{k-1}\}\\
&  =\{z\in%
%TCIMACRO{\U{211d} }%
%BeginExpansion
\mathbb{R}
%EndExpansion
^{n}|\nabla c_{k}^{T}z=0,z=f_{k-1}u_{k-1},u_{k-1}\in%
%TCIMACRO{\U{211d} }%
%BeginExpansion
\mathbb{R}
%EndExpansion
^{n}\}\\
&  =\{z\in%
%TCIMACRO{\U{211d} }%
%BeginExpansion
\mathbb{R}
%EndExpansion
^{n}|\nabla c_{k}^{T}f_{k-1}u_{k-1}=0,z=f_{k-1}u_{k-1},u_{k-1}\in%
%TCIMACRO{\U{211d} }%
%BeginExpansion
\mathbb{R}
%EndExpansion
^{n}\}.
\end{align*}
By \textit{Lemma 1}, we have%
\begin{align*}
\nabla c_{k}^{T}f_{k-1}u_{k-1}  &  =0\Leftrightarrow\\
u_{k-1}  &  =\left(  I_{n}-\frac{f_{k-1}^{T}\nabla c_{k}\nabla c_{k}%
^{T}f_{k-1}}{\delta\left(  \left\Vert f_{k-1}^{T}\nabla c_{k}\right\Vert
^{2}\right)  +\left\Vert f_{k-1}^{T}\nabla c_{k}\right\Vert ^{2}}\right)
u_{k},
\end{align*}
namely,
\[
\mathcal{V}_{1}^{k}=\mathcal{V}_{2}^{k}=\left\{  z\in%
%TCIMACRO{\U{211d} }%
%BeginExpansion
\mathbb{R}
%EndExpansion
^{n}\left\vert z=f_{k}u_{k},u_{k}\in%
%TCIMACRO{\U{211d} }%
%BeginExpansion
\mathbb{R}
%EndExpansion
^{n}\right.  \right\}
\]
where $f_{k}=f_{k-1}\left(  I_{n}-\frac{f_{k-1}^{T}\nabla c_{k}\nabla
c_{k}^{T}f_{k-1}}{\delta\left(  \left\Vert f_{k-1}^{T}\nabla c_{k}\right\Vert
^{2}\right)  +\left\Vert f_{k-1}^{T}\nabla c_{k}\right\Vert ^{2}}\right)  .$

\emph{E. Proof of Propositions in Theorem 3}

(i) Proof of \textit{Proposition 1. }In the space $%
%TCIMACRO{\U{211d} }%
%BeginExpansion
\mathbb{R}
%EndExpansion
^{n},$ the set $\mathcal{K}$ is compact iff it is bounded and closed by
Theorem 8.2 in \cite[p.41]{Morgan(2005)}. Hence, the remainder of work is to
prove that $\mathcal{K}$ is closed. Suppose, to the contrary, $\mathcal{K\ }%
$is not closed. Then there exists a sequence $x\left(  t_{n}\right)
\in\mathcal{K}\rightarrow p\notin\mathcal{K}$ with $t_{n}\rightarrow\infty.$
Whereas, $v\left(  p\right)  =\underset{t_{n}\rightarrow\infty}{\lim}v\left(
x\left(  t_{n}\right)  \right)  \leq v\left(  x_{0}\right)  $ and $c\left(
p\right)  =\underset{t_{n}\rightarrow\infty}{\lim}c\left(  x\left(
t_{n}\right)  \right)  =0$ which imply $p\in\mathcal{K}.\ $The contradiction
implies that $\mathcal{K}$ is closed. Hence, the set $\mathcal{K}$ is compact.
By (\ref{dV1}), $v\left(  x\right)  \leq v\left(  x_{0}\right)  $ with respect
to (\ref{dynamics1}), $t\geq0$. By \textit{Assumption 1}, all solutions of
(\ref{dynamics1}) satisfy $c\left(  x\right)  =0.$ Therefore, $\mathcal{K}$ is
positively invariant with respect to (\ref{dynamics1}).

(ii) Proof of \textit{Proposition 2. }Since\textit{ }$\mathcal{K}$ is compact
and positively invariant with respect to (\ref{dynamics1}), by \textit{Theorem
4.4 }(invariance principle)\textit{ }in \cite[p. 128]{Khalil (2002)}, the
solution of (\ref{dynamics1}) starting at $x_{0}$ approaches $\dot{v}\left(
x\right)  =0,$ namely $\nabla v\left(  x\right)  ^{T}f\left(  x\right)  =0.$
In addition, since (\ref{dynamics1}) becomes$\ \dot{x}=0$ in $\mathcal{S}$,
the solution approaches a constant vector $x_{l}^{\ast}\in\mathcal{S}.$

(iii) Proof of \textit{Proposition 3. }Since $\mathcal{V}_{1}\left(
x_{l}^{\ast}\right)  =\mathcal{V}_{2}\left(  x_{l}^{\ast}\right)  $ and
$x_{l}^{\ast}\in\mathcal{S}$ satisfy the following two equalities
\[
\nabla v\left(  x_{l}^{\ast}\right)  ^{T}f\left(  x_{l}^{\ast}\right)
=0,c\left(  x_{l}^{\ast}\right)  =0,
\]
there exists a $u$ such that $z=f\left(  x_{l}^{\ast}\right)  u$ for any
$z\in\mathcal{V}_{1}\left(  x_{l}^{\ast}\right)  .$ As a consequence, for any
$z\in\mathcal{V}_{1}\left(  x_{l}^{\ast}\right)  ,$ $\nabla v\left(
x_{l}^{\ast}\right)  ^{T}z=\nabla v\left(  x_{l}^{\ast}\right)  ^{T}f\left(
x_{l}^{\ast}\right)  u=0.$ There must exist $\lambda_{i}^{\ast}\in%
%TCIMACRO{\U{211d} }%
%BeginExpansion
\mathbb{R}
%EndExpansion
,$ $i=1,\cdots,m$ such that $\nabla v\left(  x_{l}^{\ast}\right)  =\sum
_{i=1}^{m}\lambda_{i}^{\ast}\nabla c_{i}\left(  x_{l}^{\ast}\right)  $.
Otherwise $\exists\bar{z}\in\mathcal{V}_{1}\left(  x_{l}^{\ast}\right)
$,$\ \nabla v\left(  x_{l}^{\ast}\right)  ^{T}\bar{z}\neq0.$ Therefore,
$x_{l}^{\ast}\in\mathcal{S}$ is a KKT point \cite[p.328]{Jorge(1999)}.
Furthermore, by Theorem 12.6 in \cite[p.345]{Jorge(1999)}, $x_{l}^{\ast}$ is a
strict local minimum if $z^{T}\nabla_{xx}L\left(  x_{l}^{\ast},\lambda^{\ast
}\right)  z>0$, for all $z\in\mathcal{V}_{1}\left(  x_{l}^{\ast}\right)
,z\neq0.$


\bigskip

\begin{thebibliography}{99}                                                                                               %


\bibitem {Tanabe(1980)}K. Tanabe, A geometric method in nonlinear
programming,\ Journal of Optimization Theory and Applications. 30(1980) 181--210.

\bibitem {Yamashita(1980)}H. Yamashita, A differential equation approach to
nonlinear programming,\ Mathematical Programming,\textit{ }18 (1980), 155--168.

\bibitem {Brown(1989)}A.A. Brown, M.C. Bartholomew-Biggs, ODE versus SQP
methods for constrained optimization,\ Journal of Optimization Theory and
Applications, 62 (1989), 371--386.

\bibitem {Zhang(1992)}S. Zhang, A.G. Constantinides, Lagrange programming
neural networks,\ IEEE Transactions on Circuits and Systems-II: Analog and
Digital Signal Processing, 39 (1992), 441--452.

\bibitem {Hou(2001)}Z.-G. Hou, A hierarchical optimization neural network for
large-scale dynamic systems,\ Automatica, 37 (2001), 1931--1940.

\bibitem {Liao(2004)}L.-Z. Liao, H. Qi, L. Qi, Neurodynamical
optimization,\ Journal of Global Optimization, 28 (2004), 175--195.

\bibitem {Barbarosou(2008)}M.P. Barbarosou, N.G. Maratos, A nonfeasible
gradient projection recurrent neural network for equality-constrained
optimization problems, IEEE Transactions on Neural Networks, 19 (2008), 1665--1677.

\bibitem {Absi(2005)}P.-A. Absi, Computation with continuous-time dynamical
systems,\ in the Grand Challenge in Non-Classical Computation International
Workshop, York, United Kingdom, 2005, Apr. 18--19.

\bibitem {Hopfield(1985)}J.J. Hopfield, D.W. Tank, Neural computation of
decisions in optimization problems,\ Biological Cybernetics, 52 (1985), 141--152.

\bibitem {Luenberger(2007)}D.G. Luenberger, Y. Ye, Linear and Nonlinear
Programming,\textit{ }third ed., Springer, Boston, 2008.

\bibitem {Lawler(1966)}E.L. Lawler, D.E. Wood, Branch-and-bound methods: a
survey,\ Operations Research, 14 (1966), 699--719.

\bibitem {Jorge(1999)}J. Nocedal, S.J. Wright, Numerical Optimization,
Springer-Verlag, New York, 1999.

\bibitem {Alberto Isdori (2003)}A. Isidori, L. Marconi, A. Serrani, Robust
Autonomous Guidance: An Internal Model-Based Approach, Springer-Verlag,
London, 2003.

\bibitem {Khalil (2002)}H.K. Khalil, Nonlinear Systems, third ed.,\textit{
}Prentice-Hall, Upper Saddle River, New York, 2002.

\bibitem {Graziano(2003)}G. Chesi, A. Garulli, A. Tesi, A. Vicino, Solving
quadratic distance problems: an LMI-Based approach, IEEE Transaction on
Automatic Control, 48 (2003), 200--212.

\bibitem {Hartley(2003)}R. Hartley, A. Zisserman, Multiple View Geometry in
Computer Vision, second ed., Cambridge University Press, Cambridge, 2003.

\bibitem {Helmke (1994)}U. Helmke, J.B. Moore, Optimization and Dynamical
Systems. Springer-Verlag, 1994.

\bibitem {Morgan(2005)}F. Morgan, Real Analysis and Applications: Including
Fourier Series and the Calculus of Variations. American Mathematical Society, 2005.
\end{thebibliography}
\end{document}